\documentclass[letterpaper, 10 pt, conference]{ieeeconf}  % Comment this line out if you need a4paper

\IEEEoverridecommandlockouts                              % This command is only needed if 
\usepackage[dvipsnames]{xcolor}
\usepackage{color}
\usepackage{acronym}
\usepackage{graphicx}
\usepackage[hidelinks]{hyperref}
\usepackage{cleveref}
\usepackage{caption}
\usepackage{subcaption}
\newcommand{\todo}[1]{}
\renewcommand{\todo}[1]{{\color{Brown}{{\tiny Todo} #1}}}

\newcommand{\density}{\sigma}

\title{\Large \bf
Sampling-free obstacle gradients and reactive planning in Neural Radiance Fields
}

\author{Michael Pantic, Cesar Cadena, Roland Siegwart, and Lionel Ott\\
\small{\texttt{\{mpantic,cesarc,rsiegwart,lioott\}@ethz.ch}}% <-this % stops a space
\thanks{All authors are with the Autonomous Systems Lab, ETH Zurich. }% <-this % stops a space
\thanks{This work was supported as a part of NCCR Digital Fabrication as well as NCCR Robotics, a National Centre of Competence in Research, funded by the Swiss National Science Foundation (grant number 51NF40\_185543).}
}

\newacro{NeRF}{Neural Radiance Field}
\newacro{SDF}{Signed Distance Function}
\newacro{MLP}{Multi-Layer Perceptron}
\newacro{BCE}{Binary Cross Entropy}
\newacro{RMP}{Riemannian Motion Policy}
\newacro{ESDF}{Euclidian Signed Distance Field}
\acrodefplural{RMP}{Riemannian Motion Policies}
\begin{document}
\maketitle
\thispagestyle{empty}
\pagestyle{empty}
%
%
%%%%%%%%%%%%%%%%%%%%%%%%%%%%%%%%%%%%%%%%%%%%%%%%%%%%%%%%%%%%%%%%%%%%%%%%%%%%%%%%
\begin{abstract}
This work investigates the use of Neural implicit representations, specifically Neural Radiance Fields (NeRF), for geometrical queries and motion planning. 
We show that by adding the capacity to infer occupancy in a radius to a pre-trained NeRF, we are effectively learning an approximation to an Euclidean Signed Distance Field (ESDF).
Using backward differentiation of the augmented network, we obtain an obstacle gradient that is integrated into an obstacle avoidance policy based on the Riemannian Motion Policies (RMP) framework. Thus, our findings allow for very fast sampling-free obstacle avoidance planning in the implicit representation.
\end{abstract}
%%%%%%%%%%%%%%%%%%%%%%%%%%%%%%%%%%%%%%%%%%%%%%%%%%%%%%%%%%%%%%%%%%%%%%%%%%%%%%%%
\section{Introduction}

In recent years a wealth of novel works on implicit map representations based on deep learning principles were published. Many of these approaches (e.g.  \cite{park2019deepsdf,lionar2021neuralblox}) are directly targeted at replacing traditional occupancy mapping systems (such as \cite{oleynikova2017voxblox, hornung2013octomap}), and thus are also fed by geometrical data such as pointclouds.
Contrastingly, the popular \acp{NeRF} \cite{mildenhall2020nerf} are trained on images and optimized for viewpoint synthesis.
While a \ac{NeRF} clearly contains \textit{some} geometrical information about the scene, the standard loss does not directly enforce the quality or spatial coherence beyond what is needed for rendering images.\\
In this paper, we investigate the use of \acp{NeRF} as an efficient map representation for motion planning, specifically obstacle avoidance which oftentimes leverages obstacle-gradient information.
In classical implicit representations the obstacle gradient can be obtained as the derivative of the \ac{SDF} \cite{oleynikova2017voxblox}. While there has been work on planning with \acp{NeRF} \cite{adamkiewicz2021}, these methods employ sampling techniques to approximate the obstacle gradient. This can become a computational bottleneck and might miss intricate features. \\
This workshop paper contributes novel insights that led to the development of sampling-free reactive collision avoidance policies based on NeRF environment representations.
First, we investigate the quality of geometric information present in the layers of a NeRF architecture and augment it to represent an SDF reliably. Second, we show how the NeRF's backwards differentiation results in an obstacle gradient. We then combine both discoveries with the RMP framework to demonstrate their applicability to obstacle avoidance.
\begin{figure}[h!]
    \centering
    \includegraphics[width=0.5\textwidth]{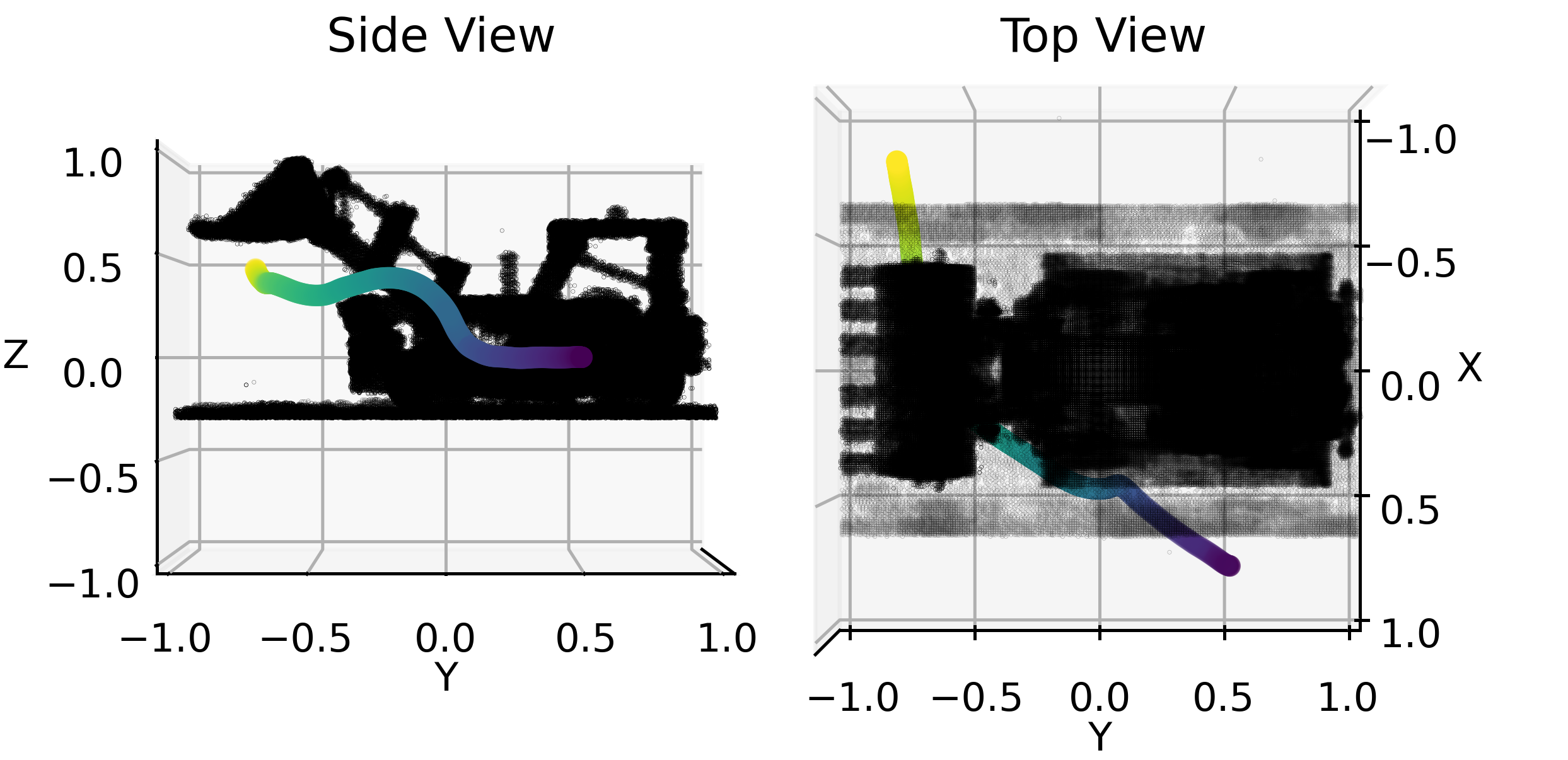}
    \caption{Example trajectory followed by an RMP moving from the start location (purple) to the goal (yellow) across the environment. Despite the noisy gradients obtained via the backwards pass, the path is smooth and does not collide with the geometry.}
    \label{fig:results_planning}
\end{figure}
\section{Method}
In the following we present the architecture and training methods of our architecture.%
\subsection{Architecture}
A common \ac{NeRF} consists of three parts -- the input positional encoding, 8 fully-connected 128-layers that output a density $\density$ plus a feature vector, and the color prediction, which uses the feature vector plus a viewpoint angle input to predict color. As color and viewpoint angle are not relevant for geometry, we remove all layers and inputs related to predicting color, and only retain the input encoding and the 8 fully-connected layers of a trained \ac{NeRF}.
The $1D$ prediction $\density$ provides a differential probability of density for a specific normalized $3D$ input coordinate $[x,y,z]$. To overcome the infinitesimal nature of $\density$, we add another fully connected layer $l_{add}$ with a single logistic-regression output $\lambda$ that is defined as
$
P(\textrm{obstacle within }r) = \lambda,
$
where $r$ is a normalized radius.
To facilitate queries with variable radii, we model $r$ as another input fed into a small intermediate layer $l_{ir}$ which is then concatenated to the add-on layer $l_{add}$. \Cref{fig:layers} visualizes the architecture. All layers are fully connected and use ReLu activation functions.
\subsection{Attachment layer}
We attach the output layer $l_{add}$ after different layers of the \ac{NeRF} \ac{MLP} during training, effectively truncating the \ac{NeRF}. Our hypothesis is that information of wider spatial extent is already accessible in the first few layers, as the full layer depth is used to predict an infinitesimal, highly localized $\density$ density value. By observing differences in training efficiency we hope to gain insight into the depth at which the add-on layer achieves the desired performance while minimizing overall model size.
\subsection{Training method}
We generate queryable occupancy data by regularly sampling $125\times10^{6}$ $\density$-densities from the \ac{NeRF} and storing points with densities above a threshold in a \texttt{kd}-tree \cite{blanco2014nanoflann}.
A training sample is generated by independently sampling $x,y,$ and $z$ coordinates from a uniform distribution $\mathcal{U}(-1,1)$, radius $r$ from a uniform distribution $\mathcal{U}(0.005,0.25)$, and the "ground-truth" occupancy classification, $\hat{y}$, by querying the \texttt{kd}-tree with the sampled $x,y,z,r$ values. 
We minimize the \ac{BCE} loss,
$\mathcal{L} = \sum_{i=0}^{n}(-\hat{y}_{i}\ log(\lambda_{i}) + (1-\hat{y}_{i})\ log(1-\lambda_{i})),$ over occupancy predictions using \textit{Adam} \cite{kingma_2014} with a batch size of $n=1000$ for $2500$ epochs.
\subsection{Motion Planning}
\acp{RMP} \cite{ratliff2018riemannian} is a framework for motion planning that provides a formulation for combining multiple policies, where each policy consists of a position- and velocity-dependent acceleration $f(x,\dot{x})$ and a dimensional weighting metric $A(x,\dot{x})$. We combine an obstacle avoidance policy with a simple goal attractor policy. For both policies we use the formulation provided in \cite{ratliff2018riemannian}\footnote{Omitted for brevity, please see Appendix "J. Collision Avoidance Controllers" in the version published at https://arxiv.org/pdf/1801.02854.pdf.}. The obstacle avoidance policy uses the map's information via obstacle gradient $\nabla d$ and distance function $d(x)$.\\
To obtain $\nabla d$ we simply use the full differentiation of the output w.r.t to the inputs as used in back-propagation, namely $\nabla d = -\left[\frac{\partial \lambda}{\partial x}, \frac{\partial \lambda}{\partial y}, \frac{\partial \lambda}{\partial z}\right],$
the obstacle distance $d(x)$ corresponds to the forward query of the network multiplied by $-1$, see \cref{sec:sdf}.
\section{Results}
For all experiments we used the \textit{Lego} dataset\footnote{Results for similar available maps, such as e.g. \textit{drums} or \textit{ship} were comparable.}, as it is widely-known and contains adequate geometry for obstacle avoidance planning.
\subsection{Optimal Attachment Layer}
\begin{figure}
    \centering
     \begin{subfigure}[b]{0.2\textwidth}
         \centering
        \includegraphics[width=\textwidth]{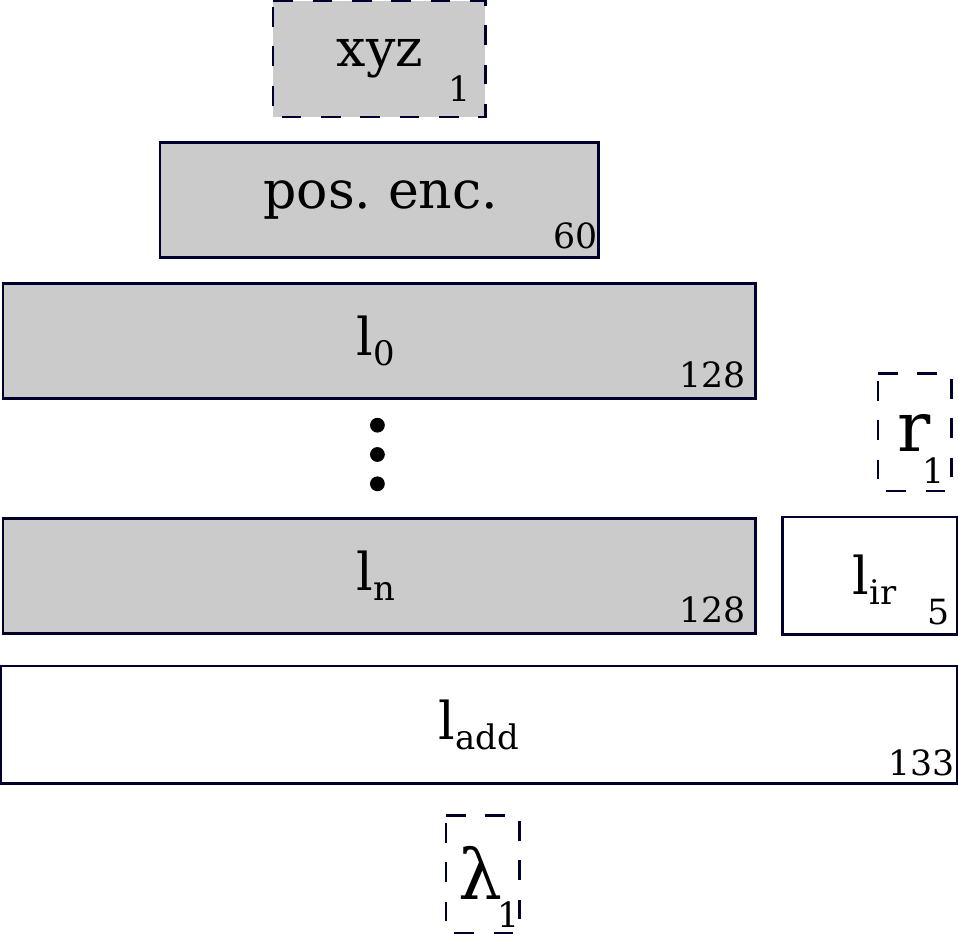}
     \end{subfigure}
     \begin{subfigure}[b]{0.25\textwidth}
         \centering
         \includegraphics[width=\textwidth]{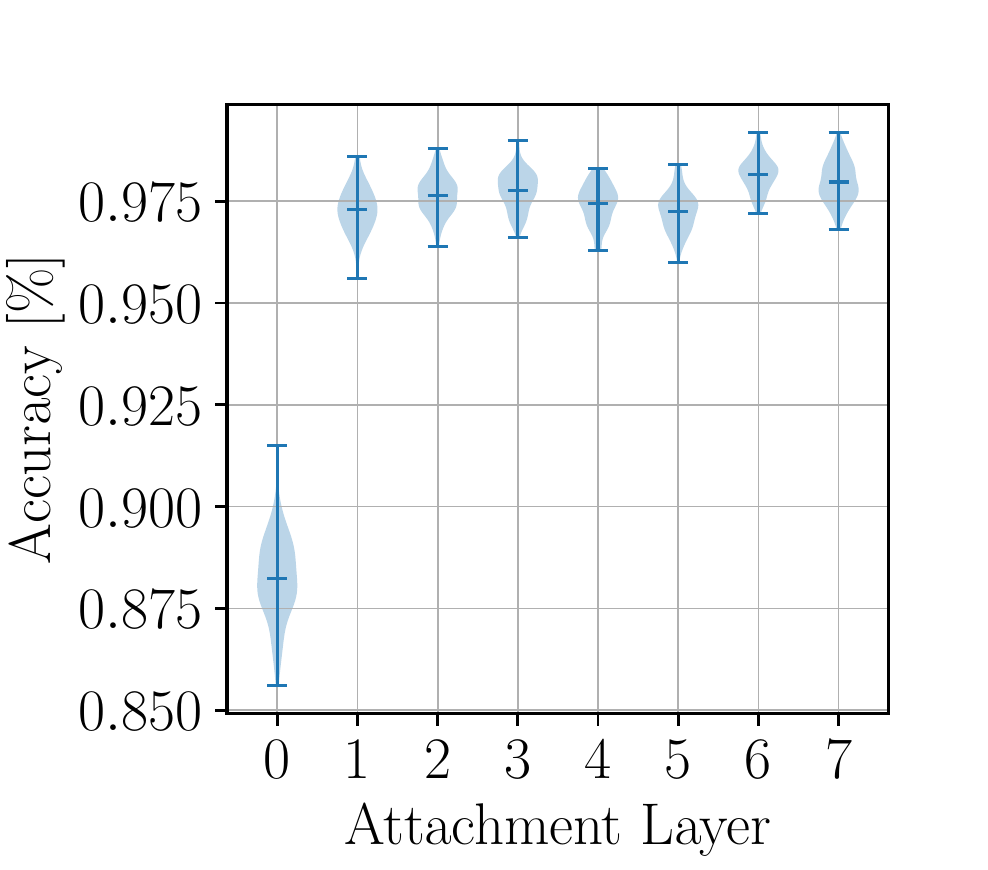}
     \end{subfigure}
    \caption{Left: Proposed architecture, frozen layers of the \ac{NeRF} are marked in grey. Right: Validation accuracy over the last 100 epochs. Attachment layer is the number of the last used pre-trained \ac{NeRF} layer.}
    \label{fig:layers}
\end{figure}

As is visible in \Cref{fig:layers}, the network obtained high accuracy for most attachment depths. Based on the achieved accuracies, we can infer that the add-on layer is able to extract meaningful information from the pre-trained \ac{NeRF} and that most of that information is present after the first few layers already.
For subsequent experiments an attachment depth of $2$ is used. While ESDF approximation showed similar results for attachment depth of $1$, the obstacle gradient improved by attaching at layer $2$.
\subsection{Occupancy queries and SDF approximation}
\label{sec:sdf}
\begin{figure}
    \centering
    \includegraphics[width=0.5\textwidth]{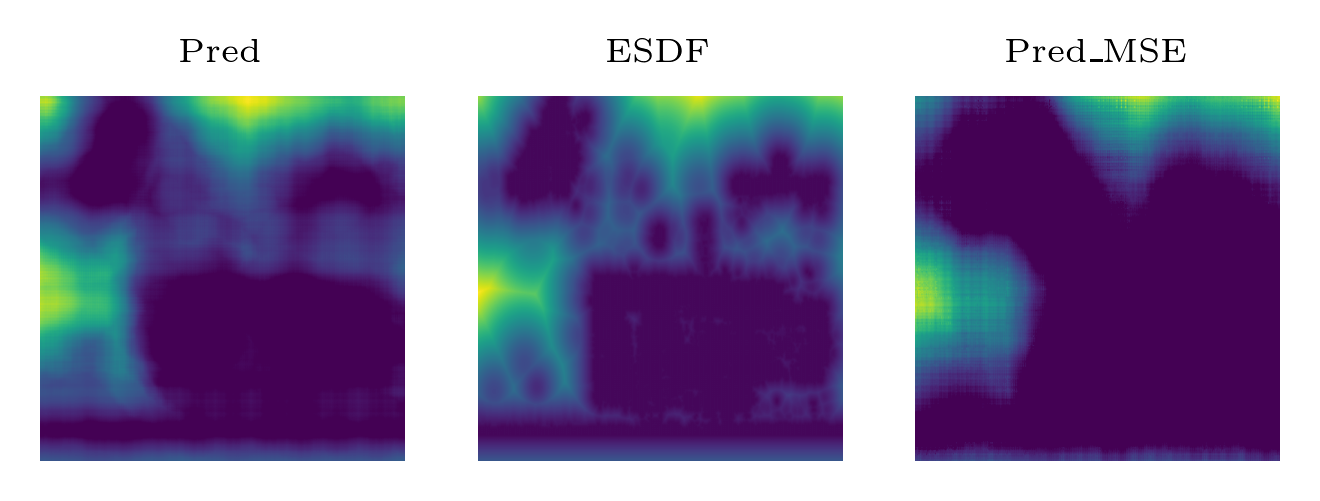}
    \caption{Non-sigmoid network output of the proposed architecture (left), ground truth ESDF (middle), and comparison against directly regressing to an ESDF using an MSE loss naively (right). Data normalized to $[0,1]$. Mean abs. error of the proposed network is $0.039$, of the naive approach $0.175$.}
    \label{fig:results_queries}
\end{figure}

While the presented architecture's main goal is to output occupancy probabilities within a certain radius, it can be used to get an approximation to an \ac{ESDF}. Taking the \textit{logit} outputs directly for a fixed radius parameter and without passing through the sigmoid function, we obtain an approximation to an \ac{ESDF}. \Cref{fig:results_queries} visualizes and compares this approximation against a ground truth \ac{ESDF}.
The resulting \ac{ESDF} is not metric, but shifted and scaled linearly. Due to the nature of the used obstacle avoidance policy, metric correctness is not needed as we can compensate this with regular parameter tuning as necessary with any \ac{RMP}.
\subsection{Motion planning}
Combining all of the aforementioned parts, we obtain a reactive motion planner that can derive the next best acceleration $\ddot{x} = f(x,\dot{x})$ in continuous space in a single forward ($d(x)$) and backward ($\nabla d(x)$) pass of the network.
Figure \ref{fig:results_planning} shows an example path planned using the combination of a goal policy and the described obstacle avoidance policy.

\section{Conclusion}
We presented a simple yet effective architecture for gradient-based planning in \acp{NeRF}. Our investigations showed that, although trained for image synthesis, the \ac{NeRF} encapsulates meaningful geometrical information that can be extracted with a simple additional layer. To our surprise most of the \ac{NeRF} layers are not needed to obtain high accuracy for geometric queries, which shows that introspection and analysis is as import in deep learning as in classical methods. Combining the discoveries of the obtained \ac{SDF} approximation and obstacle gradient, we demonstrated a sampling-free obstacle avoidance policy that only needs a single forward and backward pass per timestep. %
%%%%%%%%%%%%%%%%%%%%%%%%%%%%%%%%%%%%%%%%%%%%%%%%%%%%%%%%%%%%%%%%%%%%%%%%%%%%%%%
\bibliography{bibliography} 
\bibliographystyle{ieeetr}
\newpage
\section*{Appendix}
\begin{figure}[h!]
    \centering
    \includegraphics[width=0.5\textwidth]{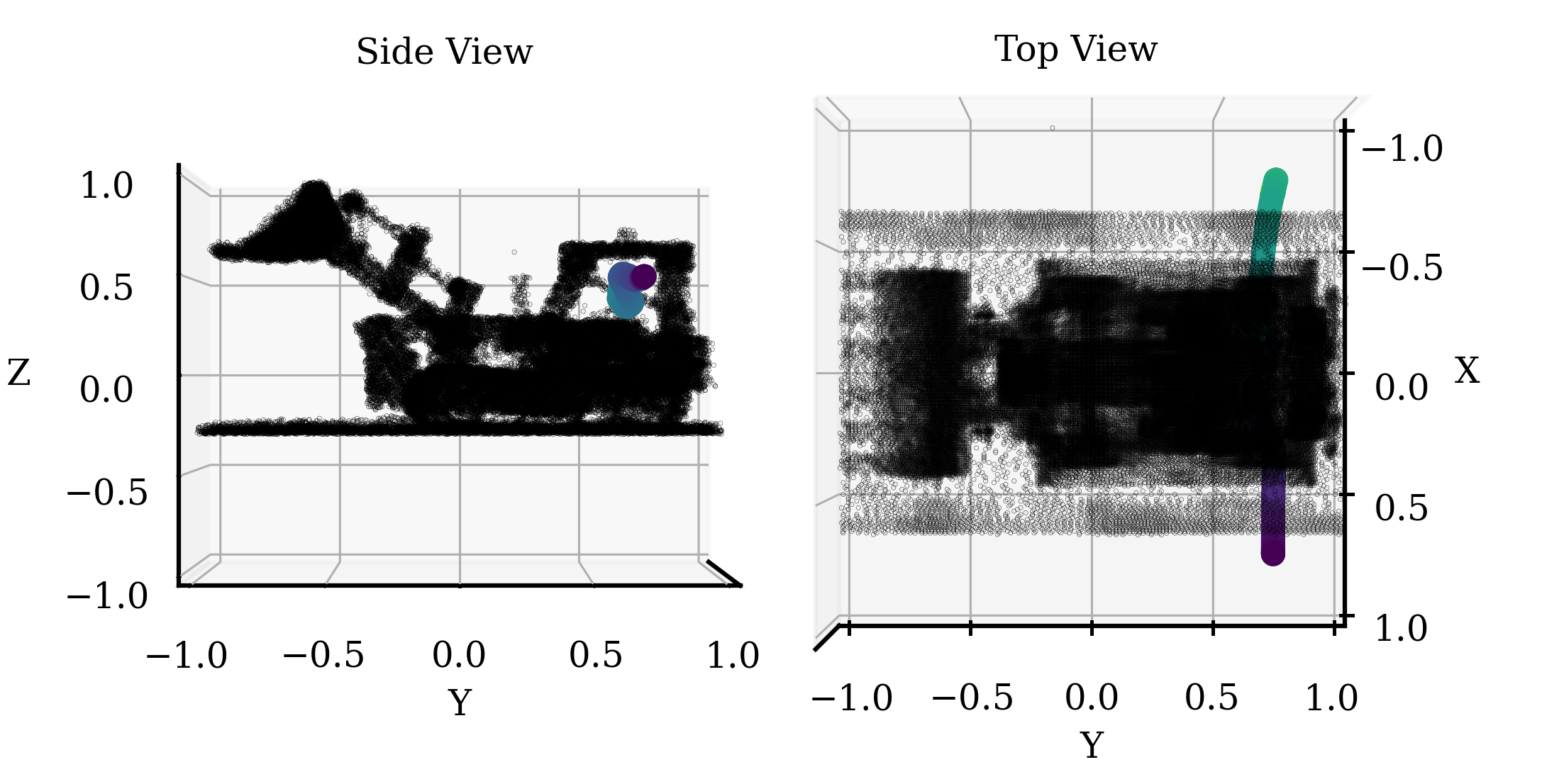}
    \caption{Example of executed plan where the planner was able to avoid intricate obstacles around the cabin of the Lego wheel loader.}
  
\end{figure}
\begin{figure}[h!]
    \centering
    \includegraphics[width=0.5\textwidth]{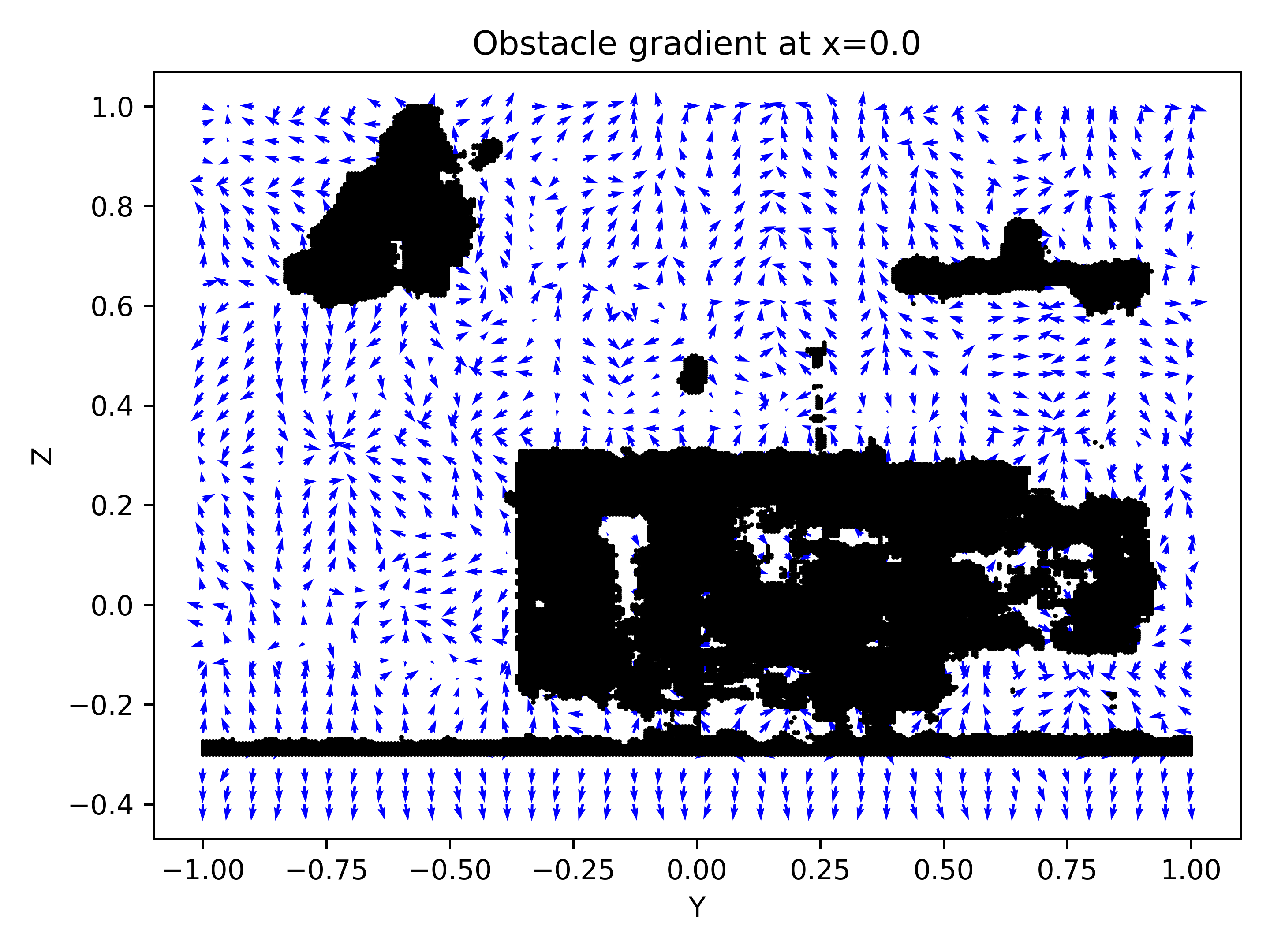}
    \caption{Example of the obstacle gradient obtained by differentiation of the network. Occupancy at $x=0$ is marked in black.}
  
\end{figure}
\end{document}